\documentclass[11pt]{article}

\usepackage[final]{acl}

\usepackage{times}
\usepackage{latexsym}

\usepackage[T1]{fontenc}

\usepackage[utf8]{inputenc}

\usepackage{microtype}

\usepackage{inconsolata}

\usepackage{algorithm}
\usepackage{algorithmic}
\usepackage{graphicx}
\usepackage{booktabs}
\usepackage{amsfonts,amssymb}
\usepackage{amsmath,bm}
\usepackage{subcaption}
\usepackage{diagbox}
\usepackage{url}
\usepackage{multirow} 
\usepackage{multicol} 
\usepackage{xcolor}
\usepackage{colortbl}
\usepackage{mathrsfs}
\usepackage{tabularx}
\usepackage{makecell}
\usepackage{arydshln}
\usepackage{bbding}
\usepackage{pifont}
\usepackage{tikz}
\usetikzlibrary{arrows.meta,positioning,shapes.geometric}

\title{ARMOR-MAD: Adaptive Routing for Heterogeneous Multi-Agent Debate in Large Language Model Reasoning}

\author{
\textbf{Fuqiang Niu}\textsuperscript{1}, 
\textbf{Bowen Zhang}\textsuperscript{2}\\
  \textsuperscript{1}School of Cyber Science and Technology, \\University of Science and Technology of China, Hefei, China\\
  \textsuperscript{2}School of Artificial Intelligence, Shenzhen Technology University, Shenzhen, China\\
  }

\begin{document}
\maketitle
\begin{abstract}
Multi-agent debate (MAD) can improve large language model reasoning, but fixed debate pipelines often waste computation and can amplify correlated errors among similar agents. We propose ARMOR-MAD, a training-free heterogeneous MAD framework that treats debate as conditional computation. ARMOR-MAD combines three components: Pre-debate Agreement Routing (PAR) decides whether independently generated Round-0 answers require debate; Early Agreement Stopping Evaluator (EASE) stops debate after convergence; and Semantic Outlier Detection (SOD) down-weights abnormal final answers during aggregation. Across MATH Level 5, GSM8K, MMLU, and MMLU-Pro, ARMOR-MAD consistently improves over fixed-round heterogeneous debate with the same model pool, reaching 65.5\%, 96.5\%, 90.0\%, and 81.5\% accuracy, respectively. The results suggest that genuine model heterogeneity and agreement-based control are both important for making MAD more accurate and efficient.
\end{abstract}

\section{Introduction}

Large language models (LLMs) have made substantial progress on mathematical reasoning, scientific question answering, and complex decision-making tasks. Yet single-model reasoning remains brittle: models often fail to self-correct without external feedback \citep{huang2024large}, chain-of-thought prompting can propagate early mistakes \citep{wei2022chain}, and confidence is not always calibrated with correctness \citep{kadavath2022language}. Multi-agent debate (MAD) addresses these weaknesses by letting multiple LLM agents generate answers, exchange rationales, critique one another, and revise their predictions \citep{du2024improving,liang2024encouraging}. Prior work reports gains in factuality, reasoning, and LLM-based evaluation \citep{du2024improving,chan2024chateval}, while broader multi-agent LLM frameworks show how conversation-based agent systems can coordinate roles and workflows \citep{he2024camel,wu2023autogen,chen2024agentverse}.

Despite these gains, many MAD systems still inherit two fragile assumptions. First, they assume that multiple agents provide independent evidence even when agents share the same backbone model or differ only through role prompts. In this setting, agents can share knowledge gaps, reasoning biases, and error modes, so debate may reinforce common mistakes instead of correcting them~\citep{zhang2024knowledge}. Second, they assume that every input benefits from the same fixed debate procedure. Fixed-round debate wastes computation on easy examples where agents already agree, and it can amplify conformity on difficult examples where agents move toward an incorrect consensus.

Recent analyses sharpen these concerns. \citet{choi2026debate} disentangle multi-agent decision making into voting and debate, showing that much of the empirical gain commonly attributed to MAD can be explained by voting rather than deliberation itself. Their theoretical analysis models debate as a martingale process over agents' belief trajectories, suggesting that debate alone does not necessarily improve expected correctness under their assumptions. Complementarily, \citet{pitre2025consensagent} identify sycophancy as a failure mode in multi-agent LLM interactions, where agents may conform to peer responses instead of critically examining them. These findings suggest that MAD should not be treated as ``more agents plus more rounds'', but as a controlled decision process whose behavior depends on agent diversity, input difficulty, and disagreement structure.

\begin{figure}[t]
\centering
\includegraphics[width=\columnwidth]{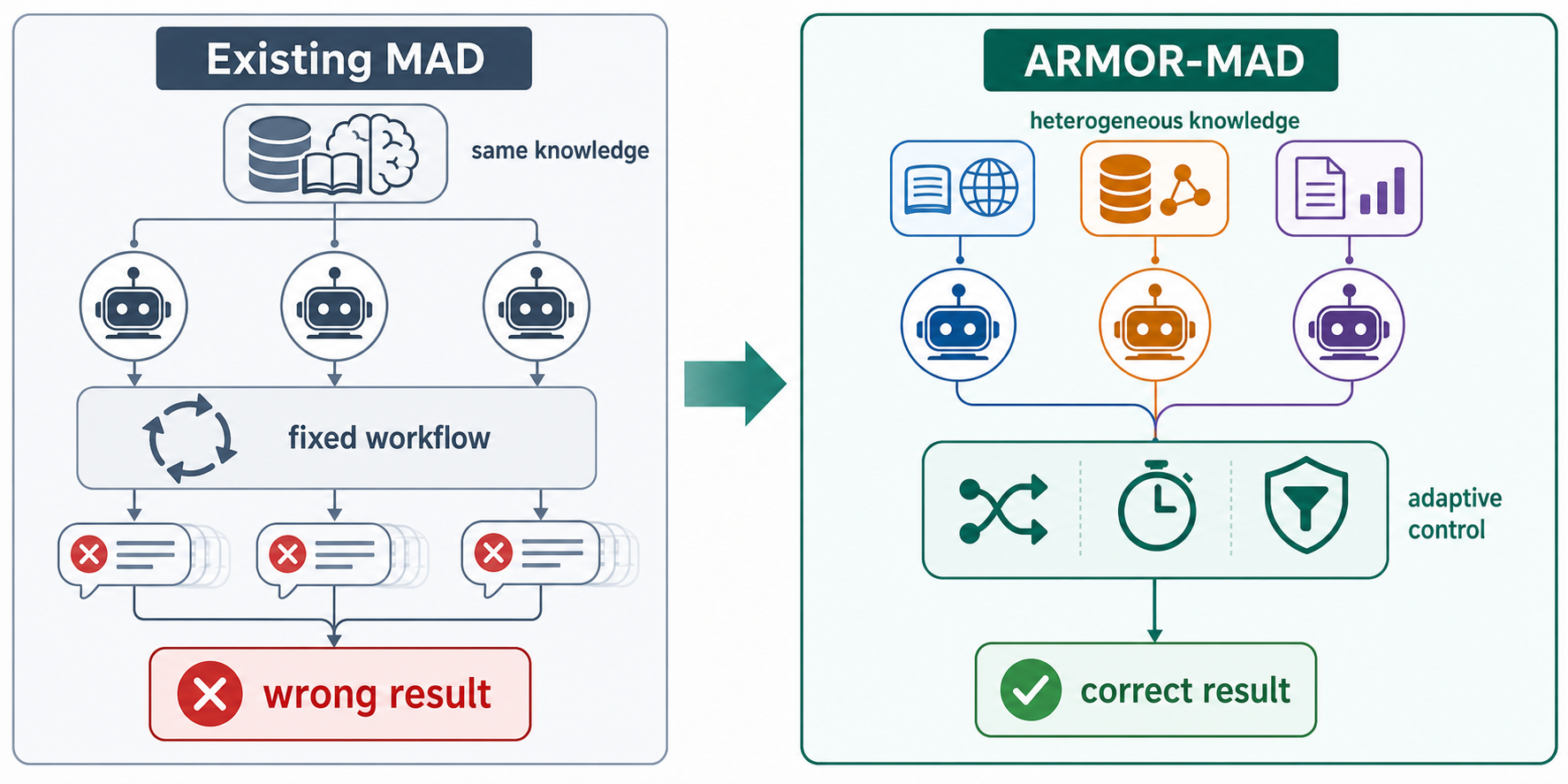}
\caption{Motivating comparison between conventional MAD and ARMOR-MAD. Conventional MAD can rely on agents with shared background knowledge and a fixed workflow, which may preserve correlated errors. ARMOR-MAD uses heterogeneous agents and adaptive control to route, stop, and aggregate reasoning according to the structure of agent agreement.}
\label{fig:intro_comparison}
\end{figure}

This view leads to two design principles. First, \textit{heterogeneity gives independent perspectives}: reducing correlated errors requires agents from genuinely different model families, not only prompt-level role variation. Second, \textit{adaptivity decides how to use those perspectives}: the system should determine when debate is necessary, when it should stop, and how final answers should be aggregated. Figure~\ref{fig:intro_comparison} summarizes the resulting contrast between fixed homogeneous debate and adaptive heterogeneous debate, motivating three coupled decisions: \textit{WHO} should participate, \textit{WHEN} debate should be triggered, and \textit{HOW} unreliable or abnormal outputs should be handled. This framing also separates agent diversity from debate control, two factors that are often conflated in prior MAD pipelines.

We propose \textbf{ARMOR-MAD}, an Adaptive Routing Multi-Agent Reasoning and Debate framework that makes these decisions in a unified pipeline. For \textit{WHO}, ARMOR-MAD uses heterogeneous LLM agents from different model families rather than prompted copies of a single backbone. For \textit{WHEN}, it introduces Pre-debate Agreement Routing (PAR), a training-free mechanism that collects independent Round-0 answers and triggers debate only when initial agreement is insufficient. Early Agreement Stopping Evaluator (EASE) then terminates debate once agents converge. For \textit{HOW}, Semantic Outlier Detection (SOD) assigns trust weights to final answers and down-weights semantically abnormal outputs. ARMOR-MAD therefore turns MAD from an indiscriminate fixed-round procedure into an adaptive pipeline over agent construction, debate invocation, stopping, and aggregation.

Our contributions are threefold. (1) We formulate multi-agent debate as an adaptive decision process governed by agent heterogeneity, debate routing, stopping, and aggregation. (2) We introduce ARMOR-MAD, a training-free heterogeneous MAD framework combining PAR, EASE, and SOD. (3) We provide empirical evidence across four reasoning and knowledge benchmarks that genuine model heterogeneity and adaptive control jointly improve performance and efficiency over fixed-round debate baselines.

\section{Related Work}
\paragraph{Multi-Agent Debate.}

Multi-agent debate has been introduced as a collective reasoning paradigm for improving the factuality and reasoning ability of large language models. 
Du et al. use multiple LLM agents to generate answers independently, exchange reasoning over several rounds, and revise predictions based on peer feedback \citep{du2024improving}. 
Liang et al. show that divergent agent perspectives can improve reasoning by increasing the diversity of intermediate solution paths \citep{liang2024encouraging}. 
Wang et al. further analyze when multi-agent discussions improve LLM reasoning, emphasizing that collaboration structure matters rather than assuming discussion is uniformly beneficial \citep{wang2024rethinking}. 
ChatEval extends debate to LLM-based evaluation, where evaluator agents deliberate over generated outputs \citep{chan2024chateval}. 
These systems establish the value of interaction, but they typically run debate as a fixed procedure instead of deciding whether each input needs debate.

Several works extend multi-agent reasoning beyond homogeneous debate and task-specific deliberation. 
ReConcile studies round-table-style discussion among diverse LLMs and uses consensus among agents to improve reasoning performance \citep{chen2024reconcile}. 
Dynamic LLM-agent networks explore adaptive collaboration structures among agents, showing that agent interaction patterns can influence final performance \citep{liu2024dynamic}. 
Agent-oriented frameworks such as CAMEL, AutoGen, and AgentVerse provide general mechanisms for role-based conversation, tool use, and multi-agent collaboration \citep{he2024camel,wu2023autogen,chen2024agentverse}. 
These studies suggest that the organization of agents matters, but they still do not directly address pre-debate routing: whether a given example should enter multi-round debate in the first place. 
ARMOR-MAD differs from prior MAD systems by treating debate as conditional computation rather than a default inference procedure.

\paragraph{Agent Heterogeneity.}

The effectiveness of multi-agent reasoning depends on whether agents provide genuinely independent perspectives. 
Standard homogeneous MAD uses multiple instances of the same backbone model, relying on sampling variation to produce useful disagreement \citep{du2024improving,liang2024encouraging}. 
However, agents from the same model family can share correlated errors, similar reasoning biases, and overlapping knowledge gaps. 
Prompt-diversified methods attempt to increase diversity by assigning agents different roles, expertise descriptions, or reasoning styles \citep{zhou2025ahmad}. 
Although role-based prompting can create surface-level behavioral differences, all agents still operate within the same underlying parameter space when they share the same backbone model.

Another line of work uses multiple model families to improve robustness. 
ReConcile studies iterative consensus formation among diverse LLMs \citep{chen2024reconcile}. 
Mixture-of-agents combines outputs from multiple proposer models through an aggregator model and improves response quality through cross-model collaboration \citep{wang2025moa}. 
However, mixture-of-agents primarily performs single-pass aggregation rather than adaptive multi-round debate \citep{wang2025moa}. 
ARMOR-MAD uses genuine model heterogeneity as the starting point, then decides whether interaction is needed from the agents' initial agreement. 
This separates two questions that prior work often entangles: how to construct diverse agents, and when their interaction should be invoked.

\paragraph{Adaptive Control.}

Adaptive control is important for multi-agent LLM systems because debate can be costly and unreliable when applied indiscriminately. 
Self-consistency improves chain-of-thought reasoning by sampling multiple reasoning paths from a single model and selecting the most consistent answer \citep{wang2023selfconsistency}. 
Structured decision fusion has also been explored in social-media stance detection, where logic-augmented frameworks combine multiple decision signals rather than relying on a single classifier prediction \citep{zhang2025logic}.
Choi et al. disentangle multi-agent decision making into majority voting and inter-agent debate, showing that much of the empirical gain commonly attributed to MAD can be explained by voting rather than deliberation itself \citep{choi2026debate}. 
Their theoretical analysis models debate as a martingale process over agents' belief trajectories, suggesting that debate alone does not necessarily improve expected correctness under their assumptions \citep{choi2026debate}. 
These findings motivate adaptive mechanisms that decide when debate should be triggered instead of applying multi-round deliberation to every input.

Consensus formation in multi-agent debate also faces robustness challenges. 
Pitre et al. identify sycophancy as a failure mode in multi-agent LLM interactions, where agents may reinforce peer responses rather than critically evaluate them \citep{pitre2025consensagent}. 
Routing-based methods address efficiency from a different perspective. 
MasRouter formulates multi-agent system routing as a learned decision problem and selects collaboration modes, agent roles, and LLM backbones through a cascaded controller \citep{yue2025masrouter}. 
While these methods improve either consensus behavior or multi-agent routing, they typically rely on interaction-time prompting strategies or learned controllers. 
ARMOR-MAD instead uses a training-free control pipeline: PAR routes examples based on Round-0 agreement, EASE stops debate once agents converge, and SOD performs semantic outlier-aware aggregation.

\section{Method}
\label{sec:method}
\paragraph{Problem Formulation.}

Given an input question $q$, a multi-agent reasoning system consists of $K$ LLM agents $\mathcal{A}=\{a_1,\ldots,a_K\}$. Each agent produces an answer $y_i$ and, optionally, a rationale $r_i$. The system returns a final answer $\hat{y}$ while controlling inference cost $C(q)$, measured by generated tokens.

Conventional MAD systems apply a fixed debate procedure to every input. In contrast, we formulate multi-agent debate as an adaptive decision process. The system must decide which agents should participate, whether debate should be triggered, when debate should stop, and how the final answer should be aggregated.

\paragraph{Overview of ARMOR-MAD.}

\begin{figure*}[t]
\centering
\includegraphics[width=\textwidth]{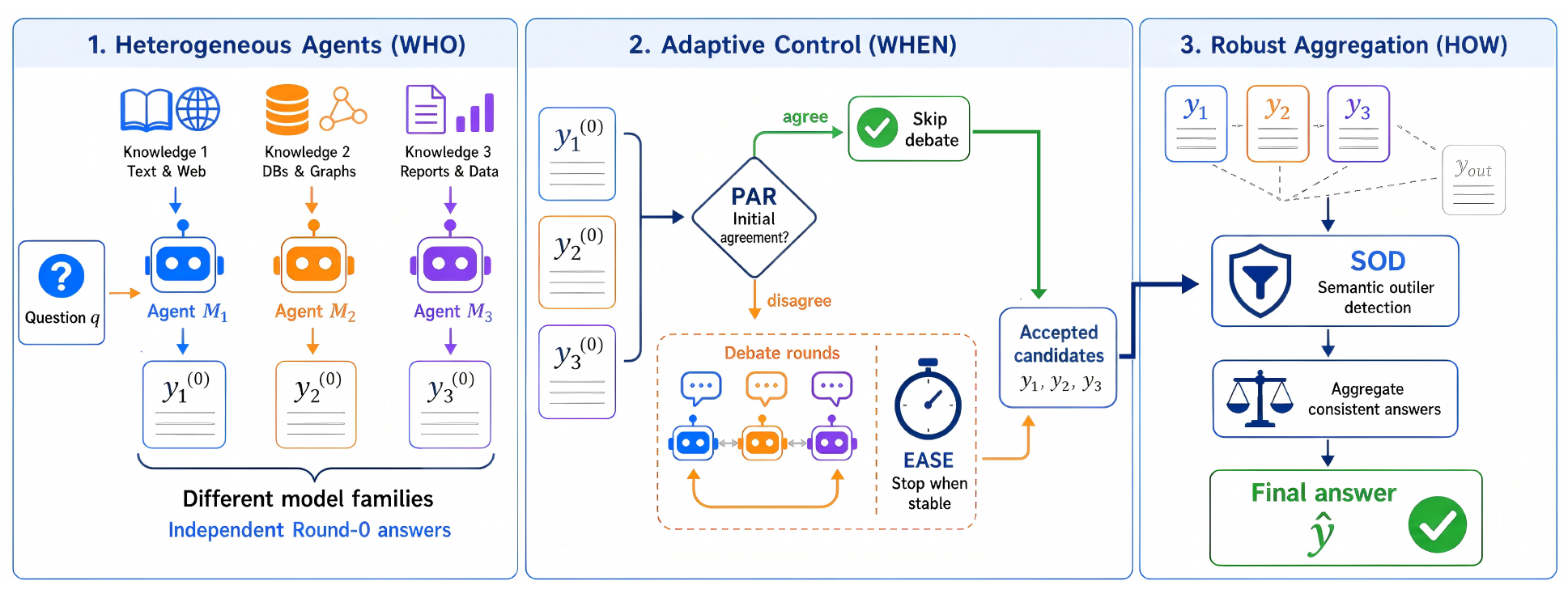}
\caption{Framework of ARMOR-MAD. Heterogeneous agents first produce independent Round-0 answers from different model families and knowledge sources. PAR decides whether the candidates already agree sufficiently; high-agreement cases skip debate, while low-agreement cases enter debate and are stopped by EASE after convergence. The resulting accepted candidates are then passed to SOD, which detects semantic outliers and aggregates consistent answers into the final prediction.}
\label{fig:method_framework}
\end{figure*}

Figure~\ref{fig:method_framework} gives an overview of ARMOR-MAD. The framework is organized around three decisions. First, for \textit{WHO}, heterogeneous agents from different model families answer the input question independently in Round 0, producing diverse candidate answers. Second, for \textit{WHEN}, PAR measures initial agreement and routes high-agreement examples directly forward, while low-agreement examples enter multi-round debate. EASE monitors the debate trajectory and stops once the agents become stable. Third, for \textit{HOW}, SOD detects semantically abnormal candidates and aggregates the consistent answers to produce the final prediction.

This design turns MAD from an indiscriminate fixed-round procedure into an adaptive pipeline. Heterogeneous agents address the agent composition problem, PAR and EASE address debate control, and SOD addresses robust aggregation. Algorithm~\ref{alg:armor} provides the corresponding procedural summary.

\paragraph{Heterogeneous Agent Construction.}

To reduce correlated errors, ARMOR-MAD uses LLMs from different model families as agents. In our implementation, the agent set consists of gpt-4o-mini, deepseek-v3, and qwen-plus. This differs from homogeneous MAD, where all agents are sampled from the same backbone model, and from prompt-diversified MAD, where agents share model parameters but receive different role or reasoning-style prompts.

The motivation is that useful debate requires agents to provide sufficiently independent perspectives. Agents from the same model family may share similar factual blind spots and reasoning biases, causing debate to reinforce common errors. In contrast, heterogeneous agents are more likely to exhibit complementary strengths and different error profiles.

\paragraph{Pre-debate Agreement Routing.}

PAR decides whether debate should be triggered before running any multi-round interaction. Given an input question $q$, each agent $a_i$ independently produces a Round-0 answer $y_i^{(0)}$. The system obtains the initial answer set:
\[
Y^{(0)}=\{y_1^{(0)},y_2^{(0)},\ldots,y_K^{(0)}\}.
\]

PAR then computes an agreement score over the Round-0 answers after task-specific answer normalization. For multiple-choice tasks, normalization maps each prediction to an option label. For mathematical reasoning tasks, normalization extracts the final numeric or symbolic answer before comparison. The agreement score is defined as the vote share of the most frequent normalized answer:
\[
\mathrm{Agree}(Y^{(0)}) = \frac{1}{K}\max_c |\{i:y_i^{(0)}=c\}|.
\]

If $\mathrm{Agree}(Y^{(0)}) \geq \tau$, ARMOR-MAD skips debate and directly aggregates the Round-0 answers. Otherwise, the system triggers multi-round debate. In our experiments, we use $\tau=0.67$, which corresponds to at least two out of three agents agreeing.

\paragraph{Early Agreement Stopping.}

For examples routed to debate, ARMOR-MAD conducts iterative interaction among agents. At each debate round $t$, every agent observes the previous-round answers and rationales from other agents, and then updates its own answer:
\[
y_i^{(t)} = a_i(q, Y^{(t-1)}).
\]

After each round, EASE computes the agreement score among current answers. If the agreement score reaches a stopping threshold $\phi$, the debate terminates early:
\[
\mathrm{Agree}(Y^{(t)}) \geq \phi.
\]

Otherwise, the debate continues until reaching the maximum number of rounds $T_{\max}$. In the main experiments, $\phi=1.0$, so EASE stops debate only when all three agents converge to the same normalized answer.

\paragraph{Semantic Outlier Detection.}

After either the Round-0 routing stage or the debate stage, ARMOR-MAD aggregates the available agent answers using Semantic Outlier Detection. Let $Y=\{y_1,\ldots,y_K\}$ denote the final answer set to be aggregated. Before aggregation, answers are normalized to task-level labels, such as option letters for multiple-choice tasks or extracted final expressions for mathematical tasks. To estimate whether an answer is semantically isolated from the rest of the group, each agent response is mapped to a TF-IDF representation $e_i$ computed over the final response text. We then compute a trust score for each answer based on its average semantic similarity to other answers:
\[
w_i = \frac{1}{K-1}\sum_{j\neq i}\cos(e_i,e_j).
\]

An answer is considered semantically abnormal if its dissimilarity exceeds a threshold:
\[
1-w_i > \lambda_{\mathrm{out}}.
\]

SOD then forms a reliability-adjusted score $\tilde{w}_i$ by setting the weight of abnormal answers to zero and keeping the original trust score for the remaining answers. The final prediction is selected from the highest-scoring normalized answer cluster:
\[
\hat{y}=\arg\max_{c}\sum_{i:y_i=c}\tilde{w}_i.
\]
If all answers are marked abnormal, ARMOR-MAD falls back to majority voting over the normalized answers. This fallback prevents the aggregation stage from discarding all available evidence in high-disagreement cases.

\begin{algorithm}[t]
\caption{ARMOR-MAD}
\label{alg:armor}
\begin{algorithmic}[1]
\REQUIRE Question $q$, agents $\mathcal{A}=\{a_1,\ldots,a_K\}$, routing threshold $\tau$, stopping threshold $\phi$, maximum rounds $T_{\max}$
\STATE Obtain independent Round-0 answers $Y^{(0)}=\{a_i(q)\}_{i=1}^{K}$
\STATE Compute initial agreement $s^{(0)}=\mathrm{Agree}(Y^{(0)})$
\IF{$s^{(0)} \geq \tau$}
    \STATE \RETURN $\mathrm{SOD}(Y^{(0)})$
\ENDIF
\FOR{$t=1$ to $T_{\max}$}
    \STATE Each agent updates its answer after observing $Y^{(t-1)}$
    \STATE Obtain updated answers $Y^{(t)}$
    \STATE Compute agreement $s^{(t)}=\mathrm{Agree}(Y^{(t)})$
    \IF{$s^{(t)} \geq \phi$}
        \STATE \textbf{break}
    \ENDIF
\ENDFOR
\STATE \RETURN $\mathrm{SOD}(Y^{(t)})$
\end{algorithmic}
\end{algorithm}

\paragraph{Computational Cost.}

ARMOR-MAD always begins with $K$ independent Round-0 model calls, one for each heterogeneous agent. The adaptive components control only the additional debate cost. Let $I(q)$ indicate whether PAR routes input $q$ to debate, and let $T(q)$ denote the number of debate rounds actually executed before EASE stops the process. The total number of agent calls is
\[
K + I(q)\cdot K\cdot T(q).
\]
By contrast, fixed-round MAD uses $K(1+T_{\max})$ calls for every input. ARMOR-MAD therefore matches fixed-round debate on low-agreement examples that require the full number of rounds, but reduces computation whenever agents agree in Round 0 or converge before $T_{\max}$. This cost model clarifies why adaptivity is useful: high-agreement examples can often be solved through heterogeneous voting and robust aggregation, while low-agreement examples still receive deliberation.

\section{Experimental Setup}

\subsection{Datasets}

We evaluate ARMOR-MAD on four benchmarks covering mathematical reasoning and knowledge-intensive question answering. 
\textbf{MATH Level 5} \citep{hendrycks2021measuring} contains high-difficulty competition-style mathematical problems requiring multi-step symbolic reasoning. 
\textbf{GSM8K} \citep{cobbe2021training} consists of grade-school mathematical word problems and is used to evaluate arithmetic and textual reasoning. 
\textbf{MMLU} \citep{hendrycks2020measuring} is a standard multiple-choice benchmark covering a broad range of academic subjects. 
\textbf{MMLU-Pro} \citep{wang2024mmlu} is a more challenging multiple-choice benchmark with more answer options and higher domain complexity. 
All methods are evaluated on exactly the same examples to ensure paired comparison. 
Detailed dataset statistics and sampling procedures are provided in Appendix~\ref{app:datasets}.

\subsection{Baselines}

We compare ARMOR-MAD with representative single-model, voting-based, debate-based, and aggregation-based baselines.
\textbf{Single-CoT} \citep{wei2022chain} uses chain-of-thought prompting as a single-agent reasoning baseline.
\textbf{Self-Consistency} \citep{wang2023selfconsistency} samples multiple reasoning paths and selects the final answer by majority vote.
\textbf{Homo-MAD} \citep{du2024improving} follows the standard multi-agent debate setting, where multiple agents debate over several rounds before final aggregation.
\textbf{D-MAD} \citep{zhou2025ahmad} is a prompt-diversified MAD baseline inspired by role-diverse heterogeneous debate, where agents are assigned different reasoning styles while sharing the same model family.
\textbf{MoA} \citep{wang2025moa} follows the mixture-of-agents paradigm, where multiple proposer outputs are synthesized by an aggregator in a single pass.
\textbf{Hetero-MAD} \citep{chen2024reconcile} is a heterogeneous debate baseline inspired by ReConcile-style multi-model discussion and consensus formation.
\textbf{Hetero-Vote-only} \citep{choi2026debate} is a voting-only diagnostic baseline motivated by recent debate-or-vote analyses, where heterogeneous agents provide initial answers without iterative debate.
Detailed model choices, prompting strategies, debate rounds, and aggregation rules for all baselines are provided in Appendix~\ref{app:baseline_details}.

\subsection{Evaluation Metrics}

Following the standard accuracy-based evaluation protocols of these benchmarks 
\citep{hendrycks2021measuring,cobbe2021training,hendrycks2020measuring,wang2024mmlu}, 
we report accuracy as the primary metric for all datasets. 
For multiple-choice benchmarks, including MMLU and MMLU-Pro, a prediction is considered correct if the extracted option matches the gold option. 
For mathematical reasoning benchmarks, including MATH Level 5 and GSM8K, we extract the final answer from the model output and compare it with the gold answer using the same normalization rules for all methods. 

\subsection{Implementation Details}

We use \texttt{gpt-4o-mini} for all single-model and homogeneous-agent baselines. 
For heterogeneous multi-agent methods, including Hetero-MAD, Hetero-Vote-only, and ARMOR-MAD, we use a fixed model pool consisting of \texttt{gpt-4o-mini}, \texttt{deepseek-v3}, and \texttt{qwen-plus}. 
Each model call is limited to 1,024 generated tokens.
For ARMOR-MAD, PAR uses an agreement threshold of $\tau=0.67$, corresponding to at least two out of three agents agreeing in Round 0. 
For routed examples, we use at most three debate rounds, and EASE stops debate once all three normalized answers converge. 
SOD uses $\lambda_{\mathrm{out}}=0.7$ as the semantic outlier threshold in the main experiments. 
Detailed prompts, baseline configurations, decoding settings, and answer extraction rules are provided in Appendix~\ref{app:implementation_details}.

\section{Experimental Results}

\subsection{Main Results}

Table~\ref{tab:main_results} reports the main results across four benchmarks. 
ARMOR-MAD obtains the highest accuracy on all four benchmarks under the main experimental setting. 
On mathematical reasoning tasks, ARMOR-MAD reaches 65.5\% on MATH Level 5 and 96.5\% on GSM8K. 
On knowledge-intensive multiple-choice tasks, ARMOR-MAD reaches 90.0\% on MMLU and 81.5\% on MMLU-Pro. 
Compared with Hetero-MAD, which uses the same heterogeneous model pool but performs fixed-round debate, ARMOR-MAD improves performance on all four datasets. 
Because Hetero-MAD uses the same model pool, this comparison isolates the contribution of adaptive routing, early stopping, and semantic aggregation beyond model heterogeneity alone.

\begin{table}[t]
\centering
\small
\resizebox{\linewidth}{!}{%
\begin{tabular}{lcccc}
\toprule
\textbf{Method} & \textbf{MATH L5} & \textbf{GSM8K} & \textbf{MMLU} & \textbf{MMLU-Pro} \\
\midrule
Single-CoT & 43.0 & 92.0 & 82.0 & 61.5 \\
Self-Consistency & 50.0 & 94.0 & 84.0 & 66.0 \\
Homo-MAD & 53.5 & 64.5 & 66.5 & 54.0 \\
D-MAD & 59.0 & 95.5 & 66.5 & 53.5 \\
Hetero-Vote-only & 50.5 & 93.0 & 87.5 & 71.5 \\
Hetero-MAD & 63.5 & 89.0 & 84.5 & 76.0 \\
MoA & 60.5 & 94.5 & 89.0 & 79.0 \\
\midrule
\textbf{ARMOR-MAD} & \textbf{65.5} & \textbf{96.5} & \textbf{90.0} & \textbf{81.5} \\
\bottomrule
\end{tabular}
}
\caption{Main accuracy results across four benchmarks.}
\label{tab:main_results}
\end{table}

The results show three patterns. 
First, homogeneous debate is unstable: Homo-MAD substantially underperforms Single-CoT on GSM8K, MMLU, and MMLU-Pro, suggesting that repeated interaction among agents from the same model family can amplify shared errors. 
Second, prompt diversity is task-dependent. 
D-MAD performs strongly on GSM8K, where reasoning-style prompts can help arithmetic word-problem solving, but it closely matches Homo-MAD on MMLU and MMLU-Pro. This pattern suggests that prompt diversity alone does not reliably remove shared knowledge blind spots in knowledge-intensive settings. 
Third, ARMOR-MAD improves over Hetero-MAD on every dataset, showing the effect of adaptive control when the model pool is fixed. 
We therefore interpret ARMOR-MAD as improving the accuracy-efficiency trade-off of heterogeneous debate rather than merely benefiting from access to stronger or more diverse base models.

\subsection{Ablation Studies}

We ablate key design choices in ARMOR-MAD. 
Table~\ref{tab:ablation} summarizes two groups of comparisons. 
Hetero-MAD removes the adaptive components and runs fixed-round heterogeneous debate. 
Hetero-Vote-only removes debate and aggregates only the initial heterogeneous answers. 
The remaining diagnostic variants isolate specific implementation choices: LLM-confidence routing replaces PAR with model-reported confidence as the routing signal, majority voting replaces SOD with simple voting, and the unstructured-output variant removes structured final-answer formatting on GSM8K.

\begin{table}[t]
\centering
\small
\setlength{\tabcolsep}{4pt}
\begin{tabular}{lcc}
\toprule
\textbf{Variant / Dataset} & \textbf{Variant} & \textbf{ARMOR} \\
\midrule
\multicolumn{3}{l}{\textit{Fixed heterogeneous debate}} \\
MATH L5  & 63.5 & \textbf{65.5} \\
GSM8K    & 89.0 & \textbf{96.5} \\
MMLU     & 84.5 & \textbf{90.0} \\
MMLU-Pro & 76.0 & \textbf{81.5} \\
\midrule
\multicolumn{3}{l}{\textit{Heterogeneous vote only}} \\
MATH L5  & 50.5 & \textbf{65.5} \\
GSM8K    & 93.0 & \textbf{96.5} \\
MMLU     & 87.5 & \textbf{90.0} \\
MMLU-Pro & 71.5 & \textbf{81.5} \\
\midrule
\multicolumn{3}{l}{\textit{Routing / aggregation variants}} \\
Conf. route & 51.5 & \textbf{65.5} \\
Maj. vote   & 89.0 & \textbf{90.0} \\
No struct.  & 89.5 & \textbf{96.5} \\
\bottomrule
\end{tabular}
\caption{Ablation results. Conf. route is evaluated on MATH L5, Maj. vote on MMLU, and No struct. on GSM8K. All numbers are accuracies.}
\label{tab:ablation}
\end{table}

The comparison with Hetero-MAD measures the effect of ARMOR-MAD's adaptive components under the same heterogeneous model pool. 
Full ARMOR-MAD improves over fixed-round heterogeneous debate on all four datasets, with the largest gains on GSM8K and MMLU. 
The comparison with Hetero-Vote-only shows that ARMOR-MAD is not merely a heterogeneous voting system: selective debate and robust aggregation improve over Round-0 voting on all four benchmarks. 
The diagnostic variants isolate individual design choices. 
Replacing PAR with LLM-confidence routing causes a large drop on MATH Level 5, suggesting that subjective model confidence is weaker than agreement among heterogeneous agents as a routing signal. 
Replacing SOD with majority voting slightly reduces MMLU accuracy, indicating that semantic aggregation provides a modest but measurable gain over simple voting. 
The GSM8K unstructured-output variant further shows that reliable final-answer formatting matters for PAR, because inconsistent numeric formats can hide agreement among agents.

\subsection{Agent Heterogeneity and Echo Chambers}

The results in Table~\ref{tab:main_results} provide evidence for two types of echo-chamber behavior. 
On GSM8K, D-MAD achieves 95.5\%, showing that prompt diversity can induce useful alternative reasoning strategies on arithmetic word problems. 
However, on MMLU and MMLU-Pro, D-MAD closely matches Homo-MAD and remains far below Hetero-MAD and ARMOR-MAD. 
This suggests that knowledge-intensive tasks can suffer from a knowledge-parametric echo chamber: agents that share the same backbone model may also share similar factual blind spots. 
In such cases, role prompts or reasoning-style prompts may not fully substitute for genuine model heterogeneity.

\subsection{Interaction Between Heterogeneity and Adaptivity}

The main results and ablations indicate that heterogeneity and adaptivity play different roles. Heterogeneity changes the evidence available to the system: agents from different model families are more likely to disagree for meaningful reasons, exposing uncertainty that homogeneous agents may hide. Adaptivity changes how the system uses that evidence: PAR treats high agreement as a signal to avoid unnecessary debate, EASE prevents additional rounds after convergence, and SOD reduces the influence of isolated final answers. The comparison between Hetero-MAD and ARMOR-MAD is therefore central because both methods use the same model pool; the gains of ARMOR-MAD come from deciding when and how to use heterogeneous opinions, rather than simply adding stronger agents.

This also explains why voting-only and fixed-debate baselines are incomplete controls. Hetero-Vote-only benefits from model diversity but cannot revise answers when agents initially disagree. Fixed Hetero-MAD allows interaction but applies it uniformly, including cases where debate is unnecessary or where additional rounds may encourage conformity. ARMOR-MAD occupies the middle ground: it preserves the efficiency of voting for high-agreement inputs while retaining debate for low-agreement cases. This supports our central view that MAD should be formulated as conditional computation over diverse agents, not as a fixed number of deliberation rounds.

\subsection{PAR Routing Behavior}

Table~\ref{tab:par_routing} reports how PAR partitions examples into skipped and debated subsets. 
Skipped examples are those for which heterogeneous agents already agree in Round 0 and therefore bypass debate. 
Across datasets, skipped examples have high accuracy, ranging from 91.2\% on MATH Level 5 to 99.4\% on GSM8K. 
This supports the main assumption behind PAR: high initial agreement among heterogeneous agents is a useful signal that debate is often unnecessary.

\begin{table}[t]
\centering
\small
\resizebox{\linewidth}{!}{%
\begin{tabular}{lcccc}
\toprule
\textbf{Metric} & \textbf{MATH L5} & \textbf{GSM8K} & \textbf{MMLU} & \textbf{MMLU-Pro} \\
\midrule
Debate Trigger Rate & 83.0 & 10.5 & 25.0 & 59.5 \\
Skip Rate & 17.0 & 89.5 & 75.0 & 40.5 \\
Skipped Accuracy & 91.2 & 99.4 & 95.3 & 96.3 \\
Debated Accuracy & 60.2 & 71.4 & 74.0 & 71.4 \\
\bottomrule
\end{tabular}
}
\caption{PAR routing behavior. Rates and accuracies are reported in percentages.}
\label{tab:par_routing}
\end{table}

The routing behavior varies substantially across datasets. 
GSM8K has the highest skip rate, indicating that most examples are already solved consistently in Round 0. 
MATH Level 5 has the highest debate-trigger rate, reflecting the difficulty of competition-style mathematical reasoning. 
MMLU and MMLU-Pro fall between these two extremes, showing that PAR adapts to both task difficulty and answer-space structure.

\subsection{Efficiency Observations}

PAR and EASE reduce computation by avoiding unnecessary debate. 
On GSM8K, ARMOR-MAD skips most examples and reduces average token usage to 2,629 tokens per example, showing the largest efficiency benefit in a high-agreement setting. 
On MATH Level 5, most examples are routed to debate, so the efficiency gain is smaller; nevertheless, ARMOR-MAD reduces average token usage from 25,725 to 23,070 compared with fixed-round Hetero-MAD. 
On MMLU-Pro, ARMOR-MAD uses 16,467 tokens on average, compared with approximately 20,000 tokens for Hetero-MAD. 
These results show that the computational benefit is input-dependent. PAR is most useful when many examples can be safely solved from Round-0 agreement, while difficult low-agreement examples still receive debate.

\subsection{Discussion}

Taken together, the results suggest that the main benefit of ARMOR-MAD is not simply adding more agents, but controlling when heterogeneous opinions should interact. 
High agreement among different model families is a strong signal that additional rounds are often redundant, as shown by the high skipped accuracy on GSM8K, MMLU, and MMLU-Pro. 
Low agreement, by contrast, indicates examples where the system should spend more computation on deliberation and robust aggregation. 
This pattern also clarifies the role of Figure~\ref{fig:intro_comparison}: the failure mode of fixed MAD is not that debate is always harmful, but that a uniform workflow cannot distinguish easy high-agreement inputs from difficult low-agreement inputs. 
ARMOR-MAD addresses this mismatch by combining heterogeneous evidence with lightweight agreement-based control.
In this sense, agreement is used not as a final answer by itself, but as a control signal for allocating deliberation where it is most likely to help.

\section{Conclusion}

We presented ARMOR-MAD, an adaptive heterogeneous multi-agent reasoning framework that treats debate as conditional computation rather than a fixed procedure. 
ARMOR-MAD combines heterogeneous agents with PAR for pre-debate routing, EASE for early stopping, and SOD for semantic outlier-aware aggregation. 
Across MATH Level 5, GSM8K, MMLU, and MMLU-Pro, ARMOR-MAD improves over fixed-round heterogeneous debate with the same model pool and achieves the best observed accuracy among evaluated methods. 
The ablations further suggest that prompt-level diversity is not a reliable substitute for genuine model heterogeneity, and that agreement-based routing can avoid unnecessary debate while maintaining strong accuracy.
This perspective makes adaptivity a first-class design choice for future MAD systems.

\section*{Limitations}

This work has several limitations. 
First, our main experiments use a fixed pool of three LLM agents, \texttt{gpt-4o-mini}, \texttt{deepseek-v3}, and \texttt{qwen-plus}. 
Although this pool is sufficient to study heterogeneity and adaptive debate, the conclusions may vary with stronger or more specialized models, larger agent sets, or different model families. 
Second, the routing and stopping rules rely on answer agreement, which is effective for short-answer and multiple-choice tasks but may be less reliable for open-ended generation, dialogue, or tasks with many semantically equivalent outputs. 
Third, SOD assumes that semantically isolated answers are more likely to be unreliable. 
This assumption can fail in ``lone expert'' cases, where a minority agent has the correct answer because it possesses domain knowledge that other agents lack. 
Fourth, our experiments are conducted on sampled subsets of four benchmarks rather than full benchmark suites, so small accuracy differences should be interpreted with statistical caution. 
Finally, ARMOR-MAD reduces unnecessary debate in high-agreement cases, but it still requires multiple model calls in Round 0; deployment in cost-sensitive settings should therefore consider both accuracy gains and API cost.

\section*{Ethical Considerations}

ARMOR-MAD is a general reasoning framework built on top of existing LLMs, and therefore inherits the risks of its underlying models. 
Heterogeneous agents may reduce some correlated errors, but they do not guarantee factual correctness, fairness, or safety. 
In high-stakes domains such as medicine, law, finance, or education, outputs from ARMOR-MAD should be treated as decision support rather than authoritative judgments, and human oversight remains necessary. 
The use of multiple commercial or closed-source models can also increase inference cost, energy consumption, and dependence on external providers. 
Because the framework routes and aggregates model outputs automatically, practitioners should monitor failure cases where agents converge on an incorrect answer or where the aggregation mechanism suppresses a correct minority response. 
We release the method as a research contribution and encourage future work on transparent routing criteria, calibrated uncertainty estimates, and stronger safeguards for sensitive applications.




\bibliography{custom}

\appendix

\section{Dataset Details}
\label{app:datasets}

We evaluate all methods on four benchmarks: MATH Level 5, GSM8K, MMLU, and MMLU-Pro. 
For the main experiments, we sample $n=200$ examples from each dataset using the same random seed, \texttt{seed}=123. 
All methods are evaluated on exactly the same sampled examples to ensure paired comparisons across systems.
All datasets are publicly available research benchmarks, and we use them only for standard evaluation following their original usage terms. 
The evaluated commercial LLMs are accessed through their official APIs under the corresponding providers' terms of service.

\paragraph{MATH Level 5.}
MATH Level 5 is the most difficult subset of the MATH benchmark \citep{hendrycks2021measuring}. 
It contains competition-style mathematical problems that require multi-step symbolic reasoning across topics such as algebra, geometry, number theory, counting and probability, prealgebra, and precalculus. 
We use Level 5 problems because they provide a challenging setting where initial disagreement among agents is common and debate may be beneficial.

\paragraph{GSM8K.}
GSM8K is a grade-school mathematical word problem benchmark \citep{cobbe2021training}. 
Problems typically require multi-step arithmetic and textual reasoning, and the final answer is usually a single integer or short numeric expression. 
This dataset is used to evaluate whether ARMOR-MAD can avoid unnecessary debate on relatively high-agreement reasoning examples.

\paragraph{MMLU.}
MMLU is a broad multiple-choice benchmark covering 57 academic subjects across STEM, humanities, social sciences, and other domains \citep{hendrycks2020measuring}. 
Each question has four answer options. 
We use MMLU to evaluate knowledge-intensive reasoning and the effect of model heterogeneity on standard multiple-choice tasks.

\paragraph{MMLU-Pro.}
MMLU-Pro is a more challenging extension of MMLU that increases the number of answer options and emphasizes more difficult reasoning-oriented questions \citep{wang2024mmlu}. 
Compared with standard MMLU, it reduces the chance-level baseline and provides a more discriminative setting for evaluating robust aggregation and adaptive debate.

\section{Baseline Details}
\label{app:baseline_details}

We provide implementation details for all baselines used in the main experiments. 
All methods are evaluated on the same sampled examples, and all outputs are processed using the same answer extraction and evaluation scripts.

\paragraph{Single-CoT.}
Single-CoT uses a single LLM with chain-of-thought prompting. 
The model is prompted to produce intermediate reasoning and then output a final answer.

\paragraph{Self-Consistency.}
Self-Consistency samples $k=5$ independent reasoning paths from the same model and aggregates the extracted final answers by majority vote.

\paragraph{Homo-MAD.}
Homo-MAD uses multiple agents instantiated from the same backbone model. 
Agents first generate independent initial answers and then participate in a fixed number of debate rounds. 
The final answers are aggregated after the last round.

\paragraph{D-MAD.}
D-MAD is a prompt-diversified MAD baseline. 
It uses agents with different reasoning-style prompts or role prompts while keeping the same underlying model family. 
This baseline is used to test whether prompt-level diversity can substitute for genuine model heterogeneity.

\paragraph{MoA.}
MoA uses multiple proposer outputs and a separate aggregation step to synthesize the final answer. 
Unlike MAD-style methods, MoA does not perform iterative inter-agent debate.

\paragraph{Hetero-MAD.}
Hetero-MAD uses agents from different model families and runs fixed-round debate. 
This baseline serves as the heterogeneous debate counterpart to ARMOR-MAD.

\paragraph{Hetero-Vote-only.}
Hetero-Vote-only collects only the initial Round-0 answers from heterogeneous agents and aggregates them without any debate. 
This diagnostic baseline is used to distinguish gains from heterogeneous voting from gains introduced by iterative debate.

\section{Implementation Details}
\label{app:implementation_details}

For single-model and homogeneous-agent baselines, we use \texttt{gpt-4o-mini}. 
For heterogeneous methods, we use \texttt{gpt-4o-mini}, \texttt{deepseek-v3}, and \texttt{qwen-plus}. 
This heterogeneous model pool is used for Hetero-MAD, Hetero-Vote-only, and ARMOR-MAD. 
Each model call is limited to 1,024 generated tokens.

All MAD-style methods use a maximum of three debate rounds. 
For ARMOR-MAD, PAR uses $\tau=0.67$ as the routing threshold. 
This corresponds to two-out-of-three agreement in Round 0. 
Examples below this threshold are routed to debate, while high-agreement examples skip debate. 
For routed examples, EASE checks agreement after each debate round and stops debate once all three normalized answers converge. 
SOD is applied as the final aggregation step with $\lambda_{\mathrm{out}}=0.7$ in the main experiments.

For mathematical datasets, models are instructed to provide an explicit final answer to reduce extraction ambiguity. 
For multiple-choice datasets, models are instructed to output one option from the given choices. 
All methods use the same answer extraction and normalization scripts. 
Dataset sampling details and baseline-specific configurations are provided in Appendix~\ref{app:datasets} and Appendix~\ref{app:baseline_details}.

\end{document}